\documentclass[twocolumn]{svjour3}       

\smartqed  
\usepackage{graphicx}
\usepackage{amssymb}
\usepackage{mathtools}
\usepackage{amsmath}
\usepackage{float}
\usepackage{epstopdf}
\usepackage{tikz}

\DeclarePairedDelimiter\floor{\lfloor}{\rfloor}
\begin{document}

\title{Automata networks for memory loss effects in the formation of linguistic conventions 
}


\author{Javier Vera  \and Eric Goles
}


\institute{J. Vera \at
              Facultad de Ingenier\'ia y Ciencias, Universidad Adolfo Ib\'a\~{n}ez, Avda. Diagonal Las Torres 2640, Pe\~{n}alol\'en, Santiago, Chile. \\
              \email{jvera@alumnos.uai.cl}           
           \and           
           E. Goles \at
              Facultad de Ingenier\'ia y Ciencias, Universidad Adolfo Ib\'a\~{n}ez, Avda. Diagonal Las Torres 2640, Pe\~{n}alol\'en, Santiago, Chile. \\
              \email{eric.chacc@uai.cl}
}

\date{Received: date / Accepted: date}

\maketitle

\begin{abstract}
This work attempts to give new theoretical insights to the absence of intermediate stages in the evolution of language. In particular, it is developed an automata networks approach to a crucial question: how a population of language users can reach agreement on a linguistic convention? To describe the appearance of sharp transitions in the self-organization of language, it is adopted an extremely simple model of (working) memory. At each time step, language users simply ``loss" part of their word-memories. Through computer simulations of low-dimensional lattices, it appear sharp transitions at critical values that depend on the size of the vicinities of the individuals. 

\end{abstract}

\section{Introduction}
\label{intro}

Contrarily to the extended view on language evolution \cite{48419,jackendoff99possibleStages} which proposes a gradual transition (through successive stages) between a ``protolanguage", a modern language minus syntax, and modern languages, recent works have been suggested the absence of intermediate stages. For instance, in \cite{FS03} it is suggested the appearance of phase transitions (scaling relations close to the \textit{Zipf's law}) in the emergence of vocabularies under \textit{least effort} constraints. 

This work attempts to give new theoretical insights to the absence of intermediate stages in the evolution of language. The startpoint is to develop a mathematical approach to a crucial question: how a population of language users can reach agreement on a linguistic convention? \cite{Steels95,Steels96,baronchelli_naming_jstat,steels2011REVIEW}. Surprisingly, language users collectively reach shared languages without any kind of central control or ``telepathy" influencing the formation of language, and only from \textit{local} conversations between few participants. The solution is based on two opposite \textit{alignment} preferences, which guide the behavior of language users by the selection of the words that give the highest chance of communicative success and the removal of the words that imply failures during communication \cite{steels2011REVIEW}. These procedures can be understand as part of the \textit{lateral inhibition} strategy \cite{Steels95}. If each convention is associated to a score measuring its amount of success, the score will decrease in the case of unsuccessful communicative interactions, and the convention will be less used. In consequence, the outcome of alignment strategies is the self-organization of agreement: the successful words will be more common, the individuals will align their own languages and there will be an increasing of the chance of successful interactions. 

To describe the appearance of sharp transitions in language formation, it is adopted an extremely simple model of (working) memory \cite{baddeley2007working}, understood as a temporal finite memory involved in on-line tasks and, specially, in language production and comprehension. At each time step, language users simply ``loss" part of their word-memories. What is more, it is hypothesized that the features of language (in particular, the consensus on a linguistic convention) emerge drastically at some critical memory loss capacity \cite{hauser1996evolution}.

The view of this work is based on a automata networks model \cite{Neumann:66,wolfram02}. Automata networks are attractive models for systems that exhibit self-organization. From extreme simplified rules of local interactions inspired in real phenomena, automata networks exhibit astonishingly rich patterns of behavior. The essential feature of the adopted framework is \textit{locality}: only from local communicative interactions, it will be described the emergence of complex language patterns.

The work proceeds by introducing basic definitions and the rules of the automata (Section 2). This is followed in Sections 3 and 4 by experiments based on an energy function that measures the amount of local agreement between individuals. Finally, a brief discussion about sharp transitions in language formation is presented. 

\section{The model}

\subsection{Basic notions}

Let $\mathcal{G}=(P,I)$ be a connected and undirected graph with vertex set $P=\{1,...,n\}$ and edge set $I$. The set $P$ represents the finite population of individuals, whereas $I$ is the set of possible interactions between individuals. A crucial element of the model is that the interactions (defined by $I$) are \textit{local}. The individual $u \in P$ participates in communicative interactions only with ``close" neighbors. To measure the degree of ``closeness" a parameter $r$ (the \textit{radius}) is introduced. The \textit{neighborhood} of radius $r$ of $u$ is the set $V^r_u=\{v \in P: 0 < d(u,v)\leqslant r\}$, where $d$ is the usual distance on $\mathcal{G}$ (the length of the shortest path between two vertices). Thus, communicative interactions occur between an individual $u \in P$ and its associated set of neighbors located on $V^r_u$.

$W$ is a finite set of words. Each individual $u \in P$ is characterized by its \textit{state} pair $(M_u,x_u)$, where $M_u$ is the memory to store words, and $x_u$ is a word of $M_u$ that $u$ conveys to the neighbors in $V^r_u$. In this context, within a communicative interaction (a vertex and its neighbors) the ``central" vertex plays the role of ``hearer", the neighbors play the role of ``speaker". Indeed, the central vertex receives the words conveyed by its neighbors. This set of conveyed words is called $W_u$, for $u \in P$. Some conveyed words are known and some of these words are unknown by the central vertex. Two sets are defined: $B_u=\{x_v \in W_u: x_v \in M_u\}$, the set of \textit{known} words, and $N_u=\{x_v \in W_u: x_v \notin M_u\}$, the set of \textit{unknown} words.

\subsection{Automata networks}

On $\mathcal{G}$, the \textbf{naming automata} is defined as the touple $\mathcal{A}=(\mathcal{G},Q,(f_u:u \in P),\phi)$, where

\begin{itemize}
\item $Q$ is the set of all possible states $\mathcal{P}(W) \times W$ ($\mathcal{P}$ denotes the set of subsets of $W$). So, the state associated to the vertex $u \in P$, $(M_u,x_u)$, is an element of $Q$ ($(M_u,x_u) \in Q$).

\item $(f_u: u \in P)$ is the set of \textit{local rules}. The naming automata $\mathcal{A}$ is \textit{uniform}, that is, each cell is associated to the same local rule. This rule takes as inputs the set $W_u$ (in particular, $B_u$ and $N_u$) and it gives as output the new state of the vertex $u$. 

\item $\phi$ is a function, the \textit{updating scheme}, that defines the order in which the vertices are updated. Traditionally, automata networks supposes the existence of a global ``clock" that establishes that all cells are updated at the same time. In this work, a \textit{fully asynchronous} scheme is considered. This updating scheme implies that at each time step one vertex is selected uniformly at random. The purpose of consider a fully asynchronous scheme arises from the typical updating order of the Naming Game. At this model, at each time step two vertices (speaker and hearer) are choosen at random. 
\end{itemize}

The configuration $X(t)$ at time step $t$ is the family $\{(M_u,x_u)\}_{u \in P}$. The vertex $u \in P$ is choosen according to the fully asynchronous scheme. The configuration at step $t+1$, $X(t+1)$, is obtained by updating through the local rule $f_u$ the state of the vertex $u$. A configuration $X'$ is a \textit{fixed point} of the dynamics if $X'(t)=X'(t+1)$ for any vertex update. 
 
\subsection{Local rules}

The local rules $(f_u:u \in P)$ are based on the concept of alignment. Suppose that at time step $t$ the vertex $u$ has been selected. $u$ and its neighbors in $V^r_u$ define a communicative interaction, in which the vertex $u$ plays the role of ``hearer", the neighbors of $V^r_u$ plays the role of ``speaker". The vertex $u$ faces with two possible actions: (1) $M_u$ is updated by adding the words of $N_u$ (\textbf{addition} action \textbf{(A)}) in order to increase the chance of future successful interactions; or (2) $M_u$ is updated by defect the words (\textbf{collapse} action \textbf{(C)}) that do not participate of successful interactions.

To measure the amount of memory loss, a third action, \textbf{forgetfulness (F)}, is introduced. Let $p \in [0,1]$ be a parameter. In simple terms, to the extent that $p$ increases, the amount of \textit{memory loss} increases. $P_u$ is the subset of $M_u \setminus \{x_u\}$\footnote{$M_u \setminus \{x_u\}$ denotes the set $M_u$ without the element $x_u$} formed by $\floor*{p(|M_u|-1)}$ words (selected at random without replacement from $M_u \setminus \{x_u\}$), where $\floor*{p(|M_u|-1)}$ means the largest integer lower than $p(|M_u|-1)$. Then, the family of local rules reads

\begin{equation}
 f_u = \left\{ \begin{array}{ll} \textrm{ if } \emptyset \neq N_u, & \left\{ \begin{array}{ll} \textbf{(F) } (M_u \setminus P_u,x_u) \\ \textbf{(A) } (M_u \cup N_u,x_u) \end{array} \right. \\ \textrm{ if }  \emptyset = N_u, & \textbf{(C) } (\{\min(B_u)\},\min(B_u))  \end{array} \right.
\end{equation}

In other words, in the case that $\emptyset \neq N_u$ the local rule acts following two steps, first, by the \textit{forgetfulness} action and, second, by the \textit{addition} action (along these two steps the set $W_i$ do not change) (see Fig. 1). 

In this paper, a particular \textbf{collapse} action is considered. Suppose that each agent is endowed with an internal total order for the set of words (equivalently, if we consider $W \subseteq \mathbb{Z}$ then the agents are endowed with the order $<$). Every agent chooses to collapse in the minimum word presented in the neighborhood. This rule represents, for example, the situation that the words differ according to their degree of relevance related to linguistic contexts \cite{Wilson04relevancetheory}. 

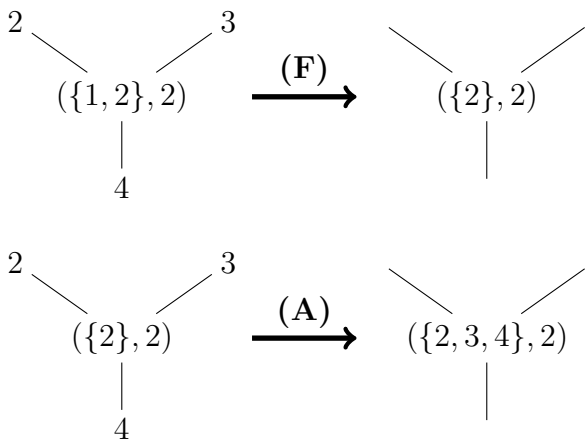
\begin{figure}
\begin{center}
\begin{tikzpicture}
  [scale=.4,auto=left,every node/.style={minimum size=0cm}] 
  [thick,scale=0.6,shorten >=0pt]
 
  \node (n1) at (3,5)  {\large $(\{1,2\},2)$}; 
   \node (n2) at (-0.5,7.5)  {\large $2$};
   \node (n3) at (6.5,7.5)  {\large  $3$};
   \node (n4) at (3,2) {\large $4$};

  \foreach \from/\to in {n1/n2,n1/n3,n1/n4}
    \draw (\from) -- (\to);
    
    
   \node (n5) at (15,5)  {\large $(\{2\},2)$}; 
   \node (n6) at (11.5,7.5)  {};
   \node (n7) at (18.5,7.5)  {};
   \node (n8) at (15,2) {};

  \foreach \from/\to in {n5/n6,n5/n7,n5/n8}
    \draw (\from) -- (\to);
    
    
   \node (n9) at (7,5)  {};
   \node (n10) at (11,5)  {};
   
   \path (n9) edge [->, line width=2pt] node[]{\large \textbf{(F)}} (n10);
   
   
   \node (n11) at (3,-3)  {\large $(\{2\},2)$}; 
   \node (n12) at (-0.5,-0.5)  {\large $2$};
   \node (n13) at (6.5,-0.5)  {\large $3$};
   \node (n14) at (3,-6) {\large $4$};

  \foreach \from/\to in {n11/n12,n11/n13,n11/n14}
    \draw (\from) -- (\to);
    
    
   \node (n15) at (15,-3)  {\large $(\{2,3,4\},2)$}; 
   \node (n16) at (11.5,-0.5)  {};
   \node (n17) at (18.5,-0.5)  {};
   \node (n18) at (15,-6) {};

  \foreach \from/\to in {n15/n16,n15/n17,n15/n18}
    \draw (\from) -- (\to);
    
    
   \node (n19) at (7,-3)  {};
   \node (n20) at (11,-3)  {};
   
   \path (n19) edge [->, line width=2pt] node[]{\large \textbf{(A)}} (n20);

\end{tikzpicture}
\end{center}
\caption{\textbf{Example of \textbf{forgetfulness (F)} and \textbf{addition (A)} actions}. $W=\{1,2,3,4\}$ is the set of words. Suppose that at some time step the vertex $u$ has been choosen. Three speakers in $V_u$ participate of the interaction. It is assumed that $\epsilon=0.5$. In the first row (F), $u$ ``forgets" the word ``1", then $(\{2\},2)$ is updated to $(\{2\},2)$. In the second row (A), $(\{2\},2)$ is updated to $(\{2,3,4\},2)$.}
\end{figure}

\section{Methods}

To explicitly describe the amount of local agreement between individuals, a function, called the ``energy", is defined (for a similar function, see \cite{Regnault20094844}). This energy-based approach arises from a physicist interpretation. The energy measures the amount of local unstability of the configuration. Large values of energy imply that the system evolves until reach \textit{ordered} configurations. At each neighborhood $V^r_u$, it is defined the function $\delta(x_u,x_v)$, $v \in V^r_u$, which is 1 in the case that $x_u=x_v$ (\textit{agreement} between the vertices $u$ and $v$), and 0 otherwise (\textit{disagreement}). Thus, it is measured the amount of local agreement of the neighborhood $\sum_{v \in V^r_u}\delta(x_u,x_v)$; summing this quantity over all vertices defines the total energy of the configuration at that time:

\begin{equation}
E(t)=-\frac{1}{n} \sum_{u \in P} \frac{1}{|V^r_u|}\sum_{v \in V^r_u}\delta(x_u,x_v)
\end{equation}

The function $E(t)$ is bounded by two extreme \textit{agreement} cases: $E(t)=0$ if all individuals convey the same word (global agreement); $E(t)=-1$ if each individual conveys a different word. The global agreement case coincides with the final absorbing state of the Naming Game, where there is one unique shared word. 

The analysis is focused on a two-dimensional periodic lattice of size $n=128^2=16384$ with Von Neumann neighborhood. The final value $E(t)$ (after $200n$ time steps or until reach $E(t)=-1$) is described for several values of $p$ and $r$: $p$ varies from 0 to 1 with an increment of $10\%$, and $r=\{1,2,3,4\}$ (respectively, 4, 12, 24 and 40 neighbors). In general, a Von Neumann neighborhood of radius $r$ supposes $2r(r+1)$ neighbors. Even though the radius $r=4$ supposes 40 neighbors, there is no a loss of \textit{locality}. Indeed,  $\frac{40}{n} \sim 2\%$ of the population of individuals.   

\section{Sharp transitions on two dimensional lattices}

\begin{figure}

  \begin{center} 
     
  \includegraphics[scale=0.4]{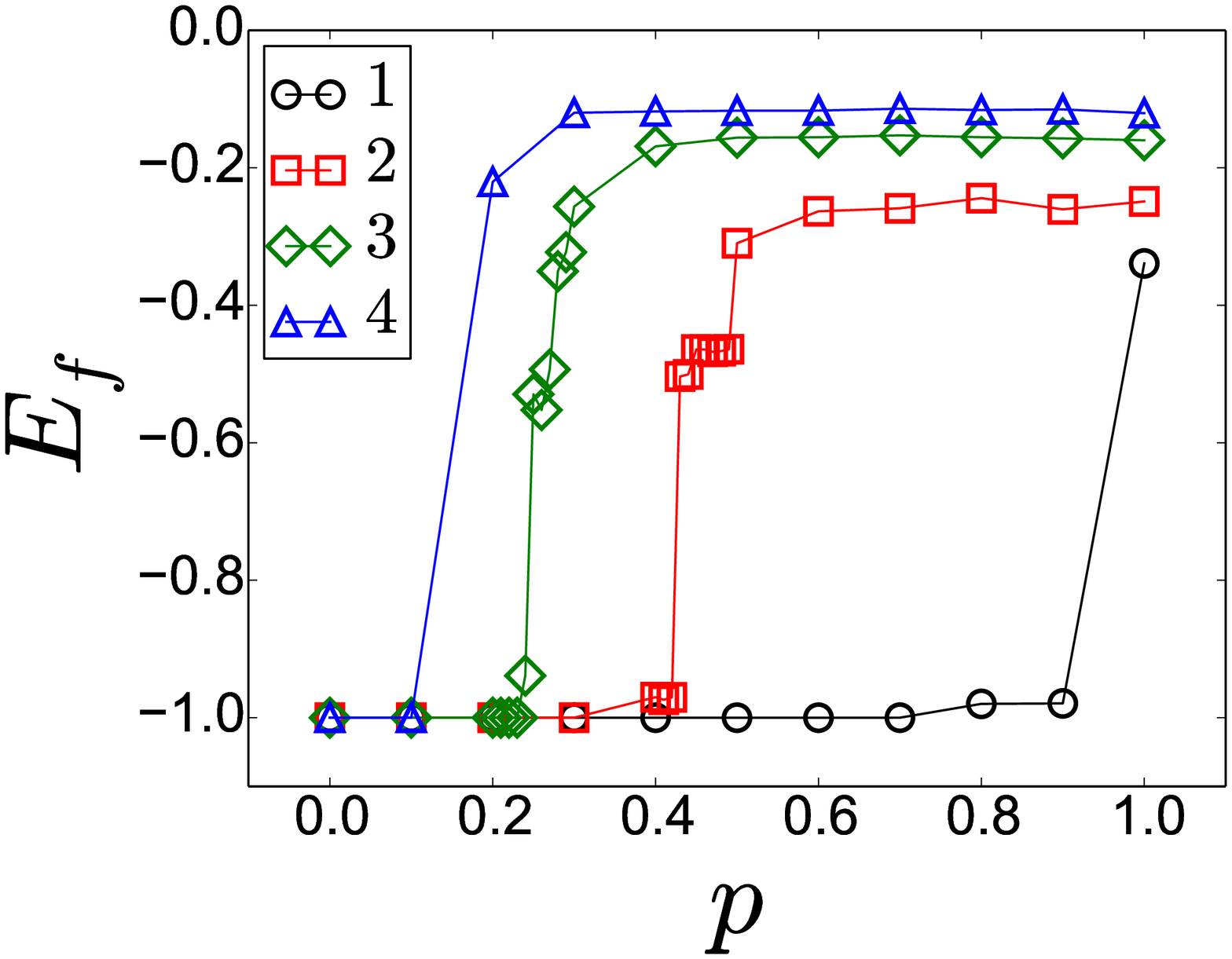}\\
  \includegraphics[scale=0.4]{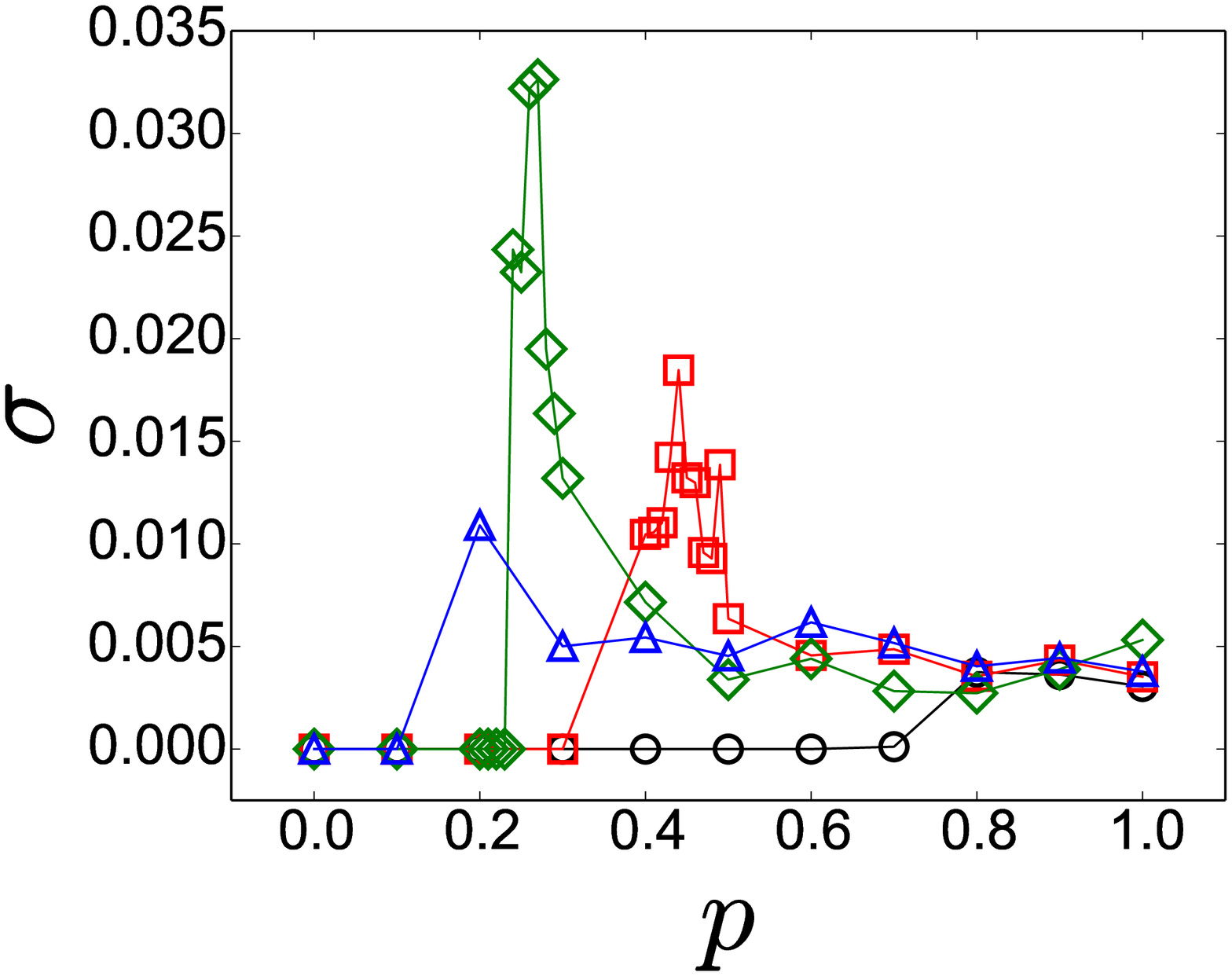}\\

  \end{center}
  \caption{\textbf{$E_f$ versus $p$ on two dimensional lattices}. \textbf{(top)} On a two dimensional lattice of size $n=128^2$, it is showed the final value of the energy function, $E_f$, versus the parameter $p$, after $200n$ steps or $E(t)$ until reach the global minimum $-1$. Averages over 100 initial conditions ($r=1,2$) and 10 initial conditions ($r=3,4$), with $|W|=n$. First, $r \in \{1,2,3,4\}$ (respectively, 4, 12, 24 and 40 neighbors) and $p$ is varied from 0 to 1 with an increment of $10\%$. With these parameters, $p^2_c \in (0.4,0.5)$, $p^3_c \in (0.2,0.3)$ and $p^4_c \in (0.1,0.2)$. New simulations run over the previous critical zones. At each critical zone, $p$ is varied with an increment of $10\%$. The critical parameter $p^r_c$, $r>1$, is then defined as the lower value of $p$ in the critical zone so that the energy function clearly does not converge to the global minimum $E(t)=-1$.  \textbf{(bottom)} Standard deviation of the data for different values of $r$.}
  
\end{figure}

Several aspects are remarkable in the behavior of $E_f$ versus $p$, as shown in Figure 2(top). For $r=1$, the dynamics reaches the configuration of global agreement ($E(t)=-1$), $p<1$. For $r=2,3,4$, the behavior of $E_f$ versus $p$ exhibits three clear domains. First, $E_f$ reaches the minimum $-1$ for $p<p^r_c$. This value depends on $r$: $p^2_c\approx 0.43$, $p^3_c \approx 0.25$, $p^4_c \approx 0.17$. In general, it is noticed that an increasing in the radius $r>1$ implies a decreasing in the critical parameter $p^r_c$. Second, a drastic change is found at $p = p_c$. The dynamics losses the convergence to the global minimum $E_f=-1$. Finally, for $p > p_c$ the dynamics seems to reach a stationary value $E_f>-1$ which increases to the extent $r$ grows. 

Standard deviation $\sigma$ of the data versus $p$, as shown in Figure 2(bottom), confirms the previous observations. Three aspects are remarkable. First, $\sigma$ takes small values in all cases. Second, for $p^r<p^r_c$, $r>1$, the standard deviation is close to 0. Third, it is observed a peak in $\sigma$ close to the critical parameters $p^r_c$, $r>1$. The values of these peaks strongly depend on the radius $r$: the more $r$ increases, the more the associated peak grows. 

\section{Discussion}

The sudden changes observed on two dimensional lattices and the peaks in standard deviation, as shown in Figure 2, and the presence of power laws (Figure 3) suggest the appearance of sharp (phase) transitions at $p=p^r_c$ for $r>1$ \cite{FS03,journals/jca/Fates09}. As it was noticed, $r=4$ exhibits the most drastic sharp transition. At the different critical forgetfulness parameters, scaling relations appear: low words (the first-ranked ones) are associated to multiple individuals, whereas several words are related to one-to-one individual-word associations. Despite of the appearance of similar slopes for $r>1$, the scaling relations for $r=4$ differs from $r=2,3$, as shown in Fig. 3. More precisely, for $r=4$ the frecuency-rank $k$ is associated to small frequencies in comparison with $r=2,3$.  

The simple approach of this paper to the individual's forgetfulness introduces a novel framework to study the influence of minimal cognitive mechanisms on the formation and evolution of languages.

Future work could involve the study of the dynamics on general topologies (for instance, random graphs), more complex cognitive mechanisms of memory capacities, or the influence of large radius $r$ (for instance, $r \sim \sqrt{n}$) on the appearance of sharp transitions.

\begin{figure}

  \begin{center} 
     
  \includegraphics[scale=0.4]{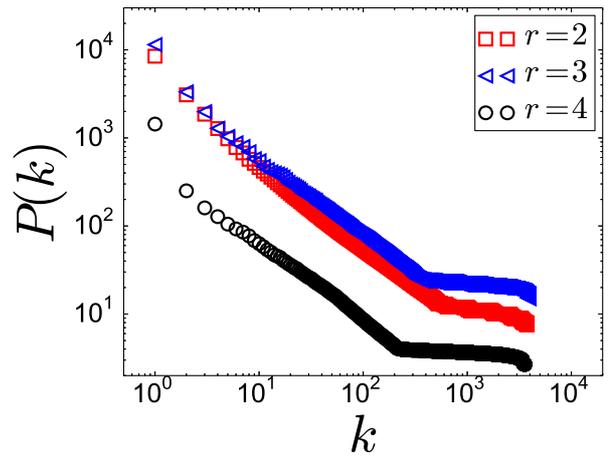}\\
 
  \end{center}
  \caption{\textbf{$P(k)$ versus $k$ on two dimensional lattices, for the critical values $p^r_c$ of $r=2$ ($p^2_c=0.43$), $r=3$ ($p^3_c=0.25$) and $r=4$ ($p^4_c=0.17$)}. On a two dimensional lattice of size $n=128 \times 128$, after $200N$ time steps it is exhibited the global distribution of the number of agents showing the $k-$ranked word, $P(k)$, versus $k$ ($\log-\log$ plot). Averages over the same initial conditions of the Figure 2.}
  
\end{figure}

\begin{acknowledgements}
The authors like to thank CONICYT-Chile under the Doctoral scholarship 21140288 (J.V) and the grants FONDECYT 11400090 (E.G), ECOS BASAL-CMM (DIM, U. Chile) (E.G), ECOS C12E05 (E.G). 
\end{acknowledgements}

\bibliographystyle{spmpsci}      
\bibliography{draft}   


\end{document}